# A Chart Generator for Shake and Bake Machine Translation

Fred Popowich

School of Computing Science, Simon Fraser University
Burnaby, British Columbia, CANADA V5A 1S6
popowich@cs.sfu.ca

**Abstract.** A generation algorithm based on an active chart parsing algorithm is introduced which can be used in conjunction with a Shake and Bake machine translation system. A concise Prolog implementation of the algorithm is provided, and some performance comparisons with a shift-reduce based algorithm are given which show the chart generator is much more efficient for generating all possible sentences from an input specification.

## 1 Introduction

Shake and Bake (S&B) is an approach to machine translation [1][14] in which generation is driven from a collection of *signs* (called a bag) where the individual signs are related by syntactic and semantic dependencies. Each sign in the bag is a feature structure containing orthographic, syntactic, and semantic information associated with a single word (or constituent) to be included in the target language sentence. This differs from some more traditional approaches to Machine Translation, as discussed in [3], in which a logical form or some explicit syntactic-semantic structure is used as the input to the generation algorithm.

The S&B generation problem is known to be NP-complete [3], but there have been suggestions for improving the efficiency of the generation process by using memoization techniques or a chart to keep track of previously encountered hypotheses [2][10], constraint networks [3][13], or heuristics [5]. If additional restrictions on the generation problem are accepted, a polynomial time algorithm can be obtained for a related generation problem [12].

Generation is used in S&B as a variation of parsing, except that a bag is consumed instead of an input string. The difference between parsing and generation is that the order of the constituents in the string constrains parsing while semantic dependencies on which constituents can be combined constrains generation.

Beaven [2] illustrated how the Cocke-Younger-Kasami (CYK) parsing algorithm could be adapted to work with a binary branching grammar (specifically, unification categorial grammar (UCG) [4]); here we will show how a more general generator, based on an active chart parser, has a straightforward realization as a Prolog program. This generator also has the potential for a variety of different search strategies aside from the breadth-first one associated with CYK. We will then examine the performance of the generator using Head-Driven Phrase Structure Grammar (HPSG) [8][9].

## 2 Shake and Bake Machine Translation

While the focus of this paper is on the generation process, it is appropriate to outline

$$s_{\langle X,...,P \rangle} \rightarrow np_{\langle X \rangle} \; vp_{\langle X,...,P \rangle}$$
$$vp_{\langle X,Y,...,P \rangle} \rightarrow v_{\langle X,Y,...,P \rangle} \; np_{\langle Y \rangle}$$
$$np_{\langle j \rangle} \rightarrow John_{\langle j \rangle}$$
$$np_{\langle m \rangle} \rightarrow Mary_{\langle m \rangle}$$
$$v_{\langle X,Y,l \rangle} \rightarrow loves_{\langle X,Y,l \rangle}$$

**Fig. 1.** Simple Grammar with Semantic Indices

briefly the translation process used in S&B since the input to the generation module is different than that found in traditional MT systems. As summarized in [10][14], the parsing of a source language sentence in S&B is performed with a unilingual grammar (and lexicon), as is the generation of a target language sentence. Instead of the parser producing an interlingual representation (to be used by a target language generator as is done in the *interlingua* approach [6]), or a language specific syntactic-semantic representation (to be subjected to language-pair specific transfer rules as is done in the *transfer* approach [6]), the goal is to produce a bag consisting of signs obtained from the leaves of the parse tree. Each of these leaves will have more information than its counterpart from the unilingual lexicon due to the unification and structure sharing that occurs as a result of a successful parse.

As a simplified example, consider the translation of the English sentence *John loves Mary* into the French sentence *Jean aime Marie*. We will assume that the source language sentence is first analyzed with the simplified English grammar and lexicon provided in Fig. 1. Instead of using an actual UCG or HPSG grammar in this paper, we can simplify the discussion by using a traditional rewrite rule notation. The last three rules in Fig. 1 would actually be lexical entries in UCG or HPSG. Each of the grammar symbols in this simplified grammar is intended to represent a feature structure, with the subscript being a list corresponding to the value of a *semantic indices* field within the feature structure. The semantic indices correspond to the arguments of the semantic relation introduced in some logical form. It is assumed that there are semantic arguments not only for the entities involved in the relation but also for the events, actions and states. In Fig. 1, we use identical uppercase letters to represent identity (structure sharing) of indices while lowercase letters are used for distinct indices.

The $X$ and $Y$ contained in the rule/entry for *loves* correspond to the semantic indices of the subject and object of the verb, with $l$ being the semantic index associated with the verb itself. So, we are saying that there is a relationship like *love(l,X,Y)*, where $l$ is the state of $X$ loving $Y$. After a successful parse, these variables in the lexical entry for *loves* will be unified or structure shared with the semantic indices found in the actual subject or object. So, in the parse of *John loves Mary*, $X$ will be unified with $j$, and $Y$ with $m$. The source language bag would thus contain the constituents [*Mary*$_{\langle m \rangle}$, *John*$_{\langle j \rangle}$, *love*$_{\langle j,m,l \rangle}$].

Given a bag of leaves obtained from the analysis of the source language sentence, a bilingual (or multilingual) lexicon is then used to obtain a corresponding bag of target language signs. Entries in the bilingual lexicon associate sets of target language signs with sets of source language signs; they equate indices from the semantic repre-sentation of the source language with ones from the semantic representation of the target language. During the transfer phase, we will assume for our simple example that the bilingual lexicon provides a one to one mapping between English words and their French counterparts and that this transfer will instantiate the semantic indices in the target language bag, giving us a bag containing the constituents [*Marie*$_{<m>}$, *Jean*$_{<j>}$, *aime*$_{<j,m,l>}$]. Note that for a sentence like *John likes Mary* which could have a translation like *Marie plaît à Jean,* the bilingual lexicon maps the single element *like* into two elements *plait* and *à*, where the order of the indices for *m* and *j* for *plaît* would be reversed from that in the source language [14].

The target language bag is given to the S&B generation algorithm which together with the unilingual grammar for the target language is used to generate the sentence. The generator determines how to combine the different constituents from the bag such that the result is compatible with the instantiated semantic indices. For example, given the target language bag above, where *j* is the semantic index of the subject and *m* is the index of the object, the generator will not be able to combine these constituents to obtain the sentence *Marie aime Jean*, but it will be able to generate *Jean aime Marie.*

## 3  A Chart Generation Algorithm

### 3.1 Chart Parsing

A chart can be viewed as a graph where the nodes correspond to positions between words in an input sentence and the edges between nodes correspond to analyses spanning substrings in the input sentence. Edges are labelled with 'dotted rules' which describe not only completed constituents (inactive edges), but also incomplete constituents (active edges). Constituents appearing to the left of the dot correspond to those that have been parsed, while those to the right of the dot (which are referred to as the 'expectations') have yet to be found. Thus, inactive edges will have no constituents yet to be found; the dot will be at the rightmost position in the rule. Fig. 2 provides part of the chart that would be created during the parsing of the sentence *John loves Mary,* assuming the grammar introduced in Fig. 2 (ignoring the semantic indices).

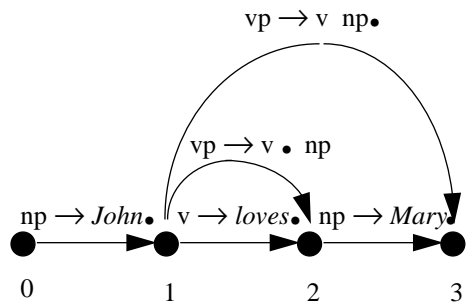

**Fig. 2.** Chart Edges and Nodes (Parsing)

During parsing, new edges can be created either by *Initialization* (also known as *scanning*), or by *Rule Invocation* (also known as *prediction*), or by Dot Movement (also known as *the fundamental rule* or *completion*). Although it is possible to describe either a top-down or bottom-up chart parser, we will only describe a bottom-up strategy (a detailed description of chart or tabular parsing from a unification-based point of view is provided in [7]). When new edges are created, they are placed into an agenda. The agenda is used to keep track of new edges until it is their turn to be added to the chart. In the simplest case, the agenda can just be a stack or queue. It is initialized from the words in the input sentence. The main parsing process consists of selecting an edge from the agenda, applying Rule Invocation and Dot Movement to the current edge in order to create new edges (which would be added to the agenda), and then adding the current edge to the chart. An edge that spans the entire input sentence corresponds to a successful parse.

In order to explicitly describe the behavior of a chart parser, we must first define initialization, rule invocation, dot movement, and the notion of a successful parse. Let us introduce the notation $<i, j, C \rightarrow \alpha \bullet \beta >$ for an edge, where $i$ is the starting position, $j$ is the ending position, $C$ is a nonterminal symbol, and $\alpha$ and $\beta$ are sequences of grammar symbols. In general, we will be using greek letters to correspond to sequences of grammar symbols, italicized lowercase letters as variables over nodes, italicized uppercase letters as variables over nonterminal symbols, and italic $w$ as a variable over terminal symbols.

**Initialization:** If word $w$ appears as the $i$th word in the input sentence then, for every grammar rule of the form $A \rightarrow w$, add edge $<i\text{-}1, i, A \rightarrow w \bullet >$ to the agenda.

**Rule Invocation:** Given inactive edge $<i, j, C \rightarrow \alpha \bullet >$, for every grammar rule of the form $A \rightarrow C \gamma$, add edge $<i, j, A \rightarrow C \bullet \gamma >$ to the agenda if it is not already in the chart or agenda.[1]

**Dot Movement:** Given active edge $<i, j, A \rightarrow \alpha \bullet C \beta >$ and inactive edge $<j, k, C \rightarrow \gamma \bullet >$, add edge $<i, k, A \rightarrow \alpha C \bullet \beta>$ to the agenda if it is not already in the chart or agenda.

**Success:** An inactive edge $<0, n, S \rightarrow \alpha \bullet >$ that spans the entire sentence of length $n$, where $S$ is the start symbol of the grammar corresponds to a successful parse of the sentence.

With these definitions in place, we explicitly describe the chart parsing algorithm in Fig. 3.

Now let us again consider the example from Fig. 2. At the time that the edge $<0,1,\text{np} \rightarrow John \bullet >$ is taken from the agenda and added to the chart, Rule Invocation is applied to this inactive edge to create the active edge $<0, 1, \text{s} \rightarrow \text{np} \bullet \text{vp}>$ which is then added to the agenda. Whenever this new active edge is then eventually processed, Dot Movement is applicable to it and the inactive edge $<1, 3, \text{vp} \rightarrow \text{v np} \bullet >$ already in the chart which results in the addition of the new edge $<0, 3, \text{s} \rightarrow \text{np vp} \bullet >$ to the agenda.

---

1. An alternative formulation of rule invocation would have the edge $<i, i, A \rightarrow \bullet C \gamma >$ added to the agenda instead. Dot movement would then be used to obtain the edge $<i, j, A \rightarrow C \bullet \gamma >$.

```
Initialization
While the agenda is not empty, remove an edge E from the agenda
    If E is an inactive edge
    Then
        If E corresponds to a Success Then Display Parse
        Attempt Rule Invocation on E
        For all active edges A in the chart, attempt Dot Movement on the pair A,E
    Otherwise /* E is an active edge */
        For all inactive edges I in the chart, attempt Dot Movement on the pair E,I
    EndIf
    Add E to the chart
```

**Fig. 3.** Chart Parsing Algorithm

This last edge will satisfy our criteria for success when it is processed, and thus an appropriate output is generated. Processing of this edge then continues, and the algorithm thus continues to look for alternative analyses for the sentence by processing subsequent edges from the agenda until the agenda is exhausted and all parses are found. Alternatively, one could terminate the algorithm after the first successful edge is found. Note that it is possible to augment the information associated with the edges so that it is easy to obtain the parse tree and/or the bag associated with the sentence.

### 3.2 Chart Generation

In chart generation, the edges of the chart still keep track of complete and incomplete constituents, but instead of the edges spanning contiguous words in an input sentence they span (not necessarily adjacent) constituents from a bag. In a chart parser, it is assumed that an edge from position $i$ to position $j$ in a sentence includes all constituents from $i$ to $j$; this is not the case with a chart generator.

Instead, we use a hypergraph as a chart with our nodes corresponding to signs from the input bag. In a hypergraph, edges connect one or more nodes. So an edge of $k$ nodes (a k-edge) in our chart corresponds to the generation of a constituent using the corresponding $k$ signs from the bag. All we need to do is assign an arbitrary numbering to each of the signs in the input bag. Our goal for generation is to obtain a hyperedge than contains all the nodes of the graph; this ensures that no sign is used more than once in the generation of a sentence.

Our notation for edges is thus changed so that instead of containing a starting and ending position, an edge contains a set of nodes connected by the hyperedge. We still use the dotted rule to label the hyperedge, so our edges are of the form $<N, C \rightarrow \alpha \cdot \beta >$ where $N$ is a set of nodes connected by the hyperedge.

Assuming the use of signs instead of atomic grammar symbols, the actual sentence generated is contained within the feature structure of the sign $C$ for a hyperedge that includes all the nodes in the chart. For a sentence $w_1, w_2, ..., w_n$, we explicitly represent it in our edges as $<\{1,2,...,n\}, C[w_1, w_2, ..., w_n] \rightarrow \alpha \cdot >$

Simply by changing the definitions for Initialization, Rule Invocation, Dot Movement, and Success, the same algorithm that was used for parsing can now be used for generation.

**Initialization (Generation):** If sign $C[w_i]$ appears as the $i^{th}$ sign in the input bag, then add edge $<\{i\}, C[w_i] \to \bullet >$ to the agenda.

**Rule Invocation (Generation):** Given inactive edge $<N, C[\omega] \to \alpha \bullet >$, for every grammar rule of the form $A \to C' \gamma$, where $C'$ unifies with $C$ under substitution $\varphi$, that is $(C)\varphi = (C')\varphi$, add edge $<N, (A[\omega] \to C \bullet \gamma)\varphi>$ to the agenda if it is not already in the chart or agenda (or more accurately, if it is not subsumed by an edge already in the chart or agenda).

**Dot Movement (Generation):** Given active edge $<M, A[\omega] \to \alpha \bullet C \beta>$ and inactive edge $<N, C'[\upsilon] \to \gamma \bullet >$, where sets $M$ and $N$ are disjoint ($M \cap N = \emptyset$), and where $C'$ unifies with $C$ under substitution $\varphi$, add edge $<M \cup N, (A[\omega\upsilon] \to \alpha C \bullet \beta)\varphi>$ to the agenda if it is not already in (or subsumed by an edge in) the chart or agenda.

**Success:** An inactive edge $<N, S[\omega] \to \alpha \bullet >$ that spans the entire bag of size $n$, $|N|=n$, where $S$ is the start symbol of the grammar corresponds to a successful generation of the sentence $\omega$.

### 3.3 An Example

To illustrate the algorithm, we will consider the generation of the sentence *Jean aime Marie* from the bag containing the signs [*Marie*$_{<m>}$, *Jean*$_{<j>}$, *aime*$_{<j,m,l>}$] with a target language grammar essentially equivalent to our simple source language grammar, except for the use of different lexical entries.

During the initialization phase the following edges will be created and added to the agenda.

$$<\{Marie_{<m>}\}, Marie_{<m>}[Marie] \to \bullet > \quad (1)$$

$$<\{Jean_{<j>}\}, Jean_{<j>}[Jean] \to \bullet > \quad (2)$$

$$<\{aime_{<j,m,l>}\}, aime_{<j,m,l>}[aime] \to \bullet > \quad (3)$$

Applying Rule Invocation to edge (1), then to edge (2), and then to (3) will result in the new edges shown in (4), (5) and (6). Notice that unification has resulted in the dotted rule from edge (6) containing more information that was originally in the grammar rule. This additional information, which specifies the indices of the arguments of the verb, came from edge (3) which came from the transfer phase of the S&B machine translation.

$$<\{Marie_{<m>}\}, np_{<m>}[Marie] \to Marie_{<m>} \bullet > \quad (4)$$

$$<\{Jean_{<j>}\}, np_{<j>}[Jean] \to Jean_{<j>} \bullet > \quad (5)$$

$$<\{aime_{<j,m,l>}\}, v_{<j,m,l>}[aime] \to aime_{<j,m,l>} \bullet > \quad (6)$$

Both edges (4) and (5) can have rule invocation applied to them, resulting in edges (7) and (8) respectively. Notice that these two new edges differ in the semantic indices associated with the vp (after the dot) that has yet to be found.

$$<\{Marie_{<m>}\},\ s_{<m,...,P>}[Marie] \rightarrow np_{<m>} \cdot vp_{<m,...,P>}> \tag{7}$$

$$<\{Jean_{<j>}\},\ s_{<j,...,P>}[Jean] \rightarrow np_{<j>} \cdot vp_{<j,...,P>}> \tag{8}$$

From edge (6), rule invocation gives us edge (9), which can then be combined with edge (4) to obtain edge (10). It would not be possible to combine edge (9) with edge (5) due to conflicting semantic indices.

$$<\{aime_{<j,m,l>}\},\ vp_{<j,m,l>}[aime] \rightarrow v_{<j,m,l>} \cdot np_{<m>}> \tag{9}$$

$$<\{Marie_{<m>}, aime_{<j,m,l>},\},\ vp_{<j,m,l>}[aime, Marie] \rightarrow v_{<j,m,l>}\ np_{<m>} \cdot > \tag{10}$$

Finally, edge (8) can combine with edge (10) to yield an edge that includes all the nodes in the chart, edge (11). Note that edge (7) would not combine with edge (10) due again to conflicting semantic indices.

$$<\{Marie_{<m>}, Jean_{<j>}, aime_{<j,m,l>}\},$$
$$s_{<j,m,l>}[Jean, aime, Marie] \rightarrow np_{<j>}\ vp_{<j,m,l>} \cdot > \tag{11}$$

## 4 Chart Generation in Prolog

### 4.1 The Chart

It is straightforward to implement edges as terms within Prolog, using the Prolog database to store the chart edges. The set of nodes associated with an edge can be stored as a bitstring; for every node $i$ contained within the set, we set the $i^{th}$ bit to 1. So the set {2,4,5} would be encoded as the bitstring 00011010 which is the integer 26, and the set {1,2,3} as 00000111 which is the integer 7. This way, we can use the logical OR operation for set union, and we can test to see if sets are disjoint by ensuring that the logical AND of their associated bitstrings is 0. The different parts of the dotted rule can be represented as separate arguments of a term. For an inactive edge of the form $<N, C\ [\omega] \rightarrow \alpha \cdot >$, we can use a term of the form

$$\text{inactive\_edge(N, C, Omega, Alpha)} \tag{12}$$

while for an active edge of the form $<M, A[\omega] \rightarrow \alpha \cdot C\ \beta>$, we could use the term

$$\text{active\_edge(M, A, Omega, Alpha, [C | Beta]).}[2] \tag{13}$$

Alternative encodings might be desirable where each edge would also contain a unique identifier number. Then, instead of having the edge explicitly store Alpha, it could instead store a list of edge numbers corresponding to the different constituents of

---

2. To improve efficiency, an alternative encoding would be preferable where inactive edges could be indexed using the **C** from the left hand side of the dotted rule, and where the active edges would be indexed using the **C** appearing immediately to the right of the dot. This could be done using the built-in indexing machanisms of Sicstus Prolog. In this way, not all edges would need to be examined when trying to apply Dot Movement (Martin Kay, personal communication).

```
cgen(Bag,Phon) :-
  init_chart_from_bag(Bag,0,AllBits), % initialize chart
  cgen_aux(AllBits,Phon).

cgen_aux(AllBits,Phon) :-
  select_next(Edge),                   % get next edge
  ( cgen(Edge,AllBits,Phon)            % succeeds when a sentence generated
  ; cgen_aux(AllBits,Phon)).           % any more edges to process?

cgen(inactive_edge(AllBits,Sym,Sentence,_), AllBits, Sentence) :-
  start_symbol(Sym).                   % success, we've generated

cgen(Edge, _, _) :- % will fail
  rule_invocation(Edge).

cgen(Edge, _, _) :-                    % will fail
  dot_movement(Edge).

dot_movement(inactive_edge(IBits,ISym,IPhrase,_)) :-
  active_edge(ActBits,ASym,APhrase,Found,[ISym|Rest]),
  IBits /\ ActBits =:= 0,              % the logical AND of the bits must be 0
  NewBits is IBits \/ ActBits,         % take the OR of the bits
  append(APhrase, IPhrase, NewPhrase),
  make_edge(NewBits, ASym, NewPhrase, [ISym|Found], Rest),
  fail.                                % failure driven loop

dot_movement(active_edge(ActBits,ASym,APhrase,Found,[ISym|Rest])) :-
  inactive_edge(IBits,ISym,IPhrase,_),
  IBits /\ ActBits =:= 0,              % the logical AND of the bits must be 0
  NewBits is IBits \/ ActBits,         % take the OR of the bits
  append(APhrase, IPhrase, NewPhrase),
  make_edge(NewBits, ASym, NewPhrase, [ISym|Found], Rest),
  fail.                                % failure driven loop

rule_invocation(inactive_edge(IBits,ISym,IPhrase,_)) :-
  rule(Sym, [ISym|Rest]),
  make_edge(IBits, Sym, IPhrase, [ISym], Rest),
  fail.                                % failure driven loop
```

**Fig. 4.** A Prolog Chart Generator

Alpha. In order to keep the implementation closely related to the description in section 3, we used the actual constituents.

**4.2 The Generator**

An outline of a chart generator, based on the chart parsing algorithm from 3.1, is pro-vided in Fig. 4. The cgen/2 predicate is given a bag of signs and it then calls the initiation process which introduces the initial edges into the agenda and returns a bitstring of 1's, AllBits, having a length corresponding to the number of constituents in the bag. Upon successful generation, cgen/2 returns a sequence of words corresponding to the generated sentence as Phon.

The cgen_aux/2 predicate selects the next edge from the agenda (adding it to the chart), and then calls cgen/3 to see if we have reached our terminating condition. It then attempts all possible applications of dot movement and rule application (using failure driven loops). Any new edges created as a result of these processes are added to the agenda by make_edge/5.

In chart parsing, it is well known that the redundancy check (where we ensure that new edges are not added to the agenda if they are already contained in the chart or agenda) is expensive. Since we are using a unification-based grammar, we would actually need a subsumption check rather than just a redundancy check to ensure that the new edge being added to the agenda is not subsumed by any edge currently in the chart or agenda. Presently, we have chosen not to implement the redundancy/subsumption check for the sake of improved performance. It could easily be added by incorporating a call to the built-in predicate subsumes_chk/2 from Sicstus Prolog within the make_edge/5 predicate.

Note that two separate clauses for dot movement are provided, one for when the current edge being processed is an active edge, another for when it is an inactive edge. While this could have been done in just one clause, two have been used for the sake of clarity. When we are processing an inactive edge (first clause), we look for all active edges in the chart such that the nodes covered by the two edges are disjoint (IBits $\wedge$ ActBits = 0), and we create a new hyperedge connecting all the nodes connected by the two original (hyper)-edges, (IBits $\vee$ ActBits). The phrase associated with the new constituent is then just the concatenation of those associated with its components.

In the clause for rule invocation, we assume that a grammar rule of the form $A \rightarrow w$, will be stored as rule(A,W), where W is a list of symbols or signs. Again, by using a failure driven loop we are able to add edges to the agenda for all possible compatible rules.

## 5 Preliminary Experiments

Using a small S&B machine translation system that was developed for HPSG [10], we replaced our HPSG shift-reduce generator with a chart generator based on the one outlined here and obtained dramatic improvement for cases in which we were interested in finding all possible sentences that could be generated. The performance of the system for generating the first (chart 1st) and all (chart all) possible sentences from bags varying in size from three to eleven elements is shown in Fig. 5. The results were obtained on a SparcStation20 running Sicstus Prolog 2.1. For each size of bag, the tests were performed for a few distinct bags, with the detailed results appearing in [11]. For the sake of comparison, the figure also includes the results (shift reduce 1st, shift reduce all) obtained from the best shift-reduce parser using memoization techniques from [10]. Note that the average time for the shift-reduce parser spent generating all possible sentences for bags of size 9, 10 and 11 (which is not included in the graph) was 3.0, 7.0 and 13.3 seconds respectively. The grammar had five rules and approximately thirty lexical entries.

It is not surprising that the chart generator shows such a marked improvement over the shift-reduce generator when exploring all alternatives, since the main concept underlying a chart generator (or parser) is to consider each hypothesis only once. The results show that the shift-reduce generator can perform marginally better than the chart generator when only the first generated sentence is required. Improvements in the chart and agenda manager, making full use of the indexing mechanisms provided by the Prolog system, should narrow (or eliminate) this margin.

In our preliminary experiments, we have only considered the performance of the system on bags from which it was possible to generate a sentence. In cases where no sentence can be generated from a bag, then both generators have to consider all possible hyphotheses (before failing). Thus, we would again expect far superior performance from the chart generator as opposed to the shift-reduce generator.

## 6 Analysis

The chart-based generation algorithm introduced in this paper is useful for generating all possible sentences from a given input bag. In cases where only one generated sentence is desired, other algorithms may be preferable. However, our algorithm is preferable in cases where translation involves the creation of several possible target language bags, each of which does not necessarily result in the successful generation of a sentence. If one is able to assume that there is only one solution generated from a bag then there is a polynomial time algorithm[12], but enough information must be transferred from the source bag to the target sign to ensure that combination is deterministic — this requires additional restrictions in the grammars and lexicons.

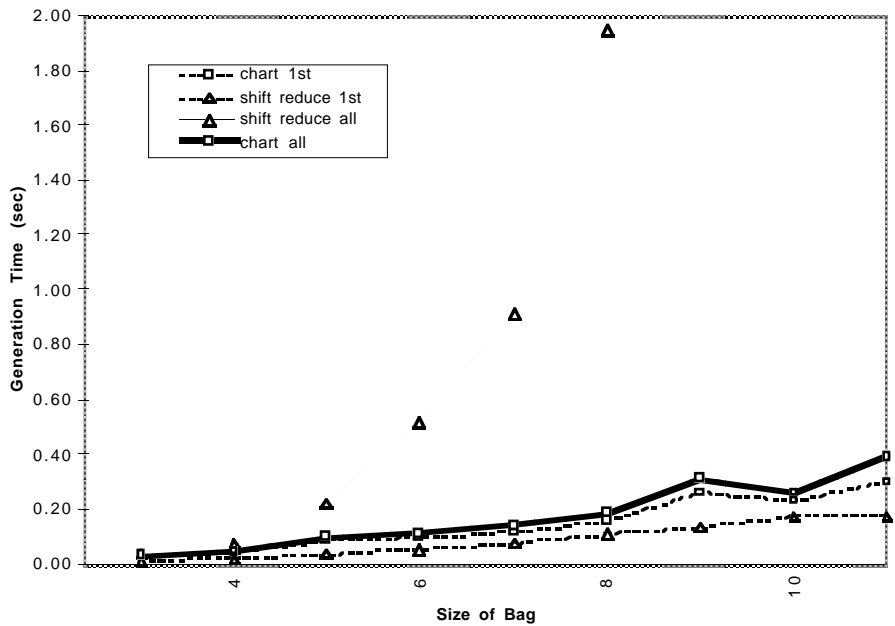

**Fig. 5.** Generation Time (secs).

Our algorithm appears to be similar to the chart generator developed independently by Trujillo and very briefly described in [13]. As Trujillo notes, it is possible to further improve the efficiency of the generation procedure by using a constraint module, exploring only the hypotheses that the constraint module deems to be possible, rather than exploring all hypotheses as is done in our algorithm. The overhead for this approach is expensive, thus making it inappropriate when dealing with uncomplicated sentences, but for "complicated sentences with several modifiers, there is a marked improvement in execution time" [13].

The algorithm presented here is more flexible than one based on the CYK algorithm [2]. The CYK algorithm incrementally builds larger and larger constituents, until one spanning all the words in a sentence is constructed. In our algorithm, there is no requirement for smaller constituents to be constructed before the larger ones. A variety of different control strategies could be used (via different agenda management techniques) to get a solution more quickly than the breadth-first strategy reflected in the CYK algorithm.

A large grammar is currently under development which will permit a proper evaluation of our algorithm on a wide range of constructions, and on a greater range of bag sizes.

## Acknowledgements

I would like to thank Dan Fass, James Devlan Nicholson and the referees for their comments and suggestions on earlier versions of this paper, and Martin Kay for discussions concerning some of the material presented in this paper. This research was supported by the Natural Sciences and Engineering Research Council of Canada by a Senior Industrial Fellowship (with TCC Communications) and a Research Grant. It was also supported by a grant from the Institute for Robotics and Intelligent Systems.

## Bibliography

1. Beaven, John L. (1992). Shake and Bake Machine Translation. In *Proceedings of the 14th International Conference on Computational Linguistics,* Nantes, France, pp. 603-609.

2. Beaven, John L. (1992). *Lexicalist Unification-Based Machine Translation.* Ph.D. thesis, Department of Artificial Intelligence, University of Edinburgh.

3. Brew, Chris (1992). Letting the Cat out of the Bag: Generation for Shake and Bake MT. In *Proceedings of the 14th International Conference on Computational Linguistics,* Nantes, France, pp. 610-616.

4. Calder, Jo, Ewan Klein, and Henk Zeevat (1988). Unification Categorial Grammar: A Concise Extendable Grammar for Natural Language Processing. In *Proceedings of the 12th International Conference on Computational Linguistics*, Budapest, Hungary, pp. 83-86.

5. Chen, Hsi-Hsi and Yue-Shi Lee (1994). A Corrective Training Algorithm for Adaptive Learning in Bag Generation. In *International Conference on New Methods in Language Processing (NeMLaP),* UMIST, Manchester, UK, pp. 248-254.

6. Hutchins. W.J. and Harry L. Somers (1992). *An Introduction to Machine Translation*. London, Academic Press.

7. Pereira, Fernando C.N. and Stuart M. Shieber (1987). *Prolog and Natural Language Analysis*, CSLI Lecture Notes, University of Chicago Press.

8. Pollard, Carl, and Ivan Sag (1987). *Information-Based Syntax and Semantics, Volume 1: Fundamentals*. CSLI, Stanford University, CA.

9. Pollard, Carl, and Ivan Sag (1994). *Head-Driven Phrase Structure Grammar.* Centre for the Study of Language and Information, Stanford University, CA.

10. Popowich, Fred (1995). Improving the Efficiency of a Generation Algorithm for Shake and Bake Machine Translation Using Head-Driven Phrase Structure Grammar. In *Proceedings of the Fifth International Workshop on Natural Language Understanding and Logic Programming,* Lisbon, Portugal.

11. Popowich, Fred (1995). A Chart Generator for Shake and Bake Machine Translation. Technical Report CMPT TR 95-08, School of Computing Science, Simon Fraser University, July 1995.

12. Poznanski, Victor, John L. Beaven and Pete Whitelock (1995). An Efficient Generation Algorithm for Lexicalist MT. In *Proceedings of the 33rd Annual Meeting of the Association for Computational Linguistics,* Cambridge, MA.

13. Trujillo, Arturo (1995). Machine Translation with the ACQUILEX LKB. Working Paper, Computer Laboratory, University of Cambridge, July 1, 1995.

14. Whitelock, Pete (1994). Shake and Bake Translation. In C.J. Rupp, M.A. Rosner and R.L. Johnson (eds.) *Constraints, Language and Computation,* London, Academic Press, pp. 339-359.